# A WAVELET, AR AND SVM BASED HYBRID METHOD FOR SHORT-TERM WIND SPEED PREDICTION

G.V. DRISYA[a] AND K. SATHEESH KUMAR[b1]

[ab]Department of Futures Studies, University of Kerala, Kariyavattom, Thiruvananthapuram, Kerala, India

## ABSTRACT

Wind speed modelling and prediction has been gaining importance because of its significant roles in various stages of wind energy management. In this paper, we propose a hybrid model, based on wavelet transform to improve the accuracy of the short-term forecast. The wind speed time series are split into various frequency components using waveletdecomposition technique, and each frequency components are modelled separately. Since the components associated withthe high- frequency range shows stochastic nature, we modelled them with autoregressive (AR) method and rest of low-frequencycomponents modelled with support vector machine (SVM). The results of the hybrid method show a promisingimprovement in accuracy of wind speed prediction compared to that of stand-alone AR or SVM model.

As a renewable source of energy, the wind is considered as a key solution to reach a sustainable future. At the end of 2016, worldwide total installed wind power capacity is 486.749 GW and it has steadily been increasing, and by 2030 is expected to reach 2,110 GW supplying 20% of global electricity, (GWEC 2014). In a local scenario, India is in the fourth position in the global wind industry with a total installed capacity of31 GW by the end of 2016 and also expecting 639 GW of installed capacity by 2020, (GWECIndia 2017).Any improvement in wind speed modelling and prediction contribute the cost-effectiveness and reliability ofwind energy technology and worldwidemany research organisations are working on its different aspects.The International Energy Agency Wind (IEA Wind), pull together the countries that are members in it, forcombined activities to address the research needs of wind energy management and thereby benefiting the entirewind energy community. To direct these efforts on opportunities in strategic areas like resource, design,operation, integration, and social and environmental impacts, IEA Wind identified and documented severalrelevant research topics, being characterising the wind as one of them, (IEA 2013). Modelling wind speeddynamics addresses the wind resource characterisation and it deals with the research needs associated withsite optimisation, operation of the power plant and wind turbine performance &output prediction. While thestudies on the long-term wind characteristics of the location are critical for good site selection, short-term andmedium-term wind prediction contribute to the other needs to reduce operation and maintenance cost. Sincewind speed oscillations are is highly variable in nature, integrating wind power into the traditional transmissiongrid presents some challenges on different aspects of wind energy management, including transmissionplanning and power quality, (Georgilakis 2008). Prior knowledge of wind power, at different time scales ofinterest, can contribute in these areas so as utility operators can plan transmission and marketing strategies.

In general, wind speed forecasting methods can be classified into two: physical models such as NumericalWeather Prediction model (NWP) and time series models such as Autoregressive (AR) model, (Soman et al.2010). Physical models are location specific, mathematically complex, computationally intensive and timeconsuming making themunreliable for short and medium term wind speed prediction, (Potter &Negnevitsky2006, Candy et al. 2009). Time series analysis methodologies are generally considered as suitable for modellingany natural system from its historical data and hence are fit to derive a good description of highly intermittentwind speed data. The use of statistical methods in explaining the random variation in wind speed datais available in the literature. The capability of Gaussian distribution to characterise wind over time explainedby, (Brown et al. 1984) was the first attempt in this direction. Assuming wind as a stochastic process manyother statistical models such as AR, ARMA, ARIMA and some probability distribution based models are alsoexperimented to explain wind speed oscillations, (Kamal & Jafri 1997, Cadenas& Rivera 2007, Kavasseri & Seetharaman 2009, Hennessey Jr 1977, Celik 2004, Mathew et al. 2011, Jiang et al. 2013). Investigationsbased on data mining models such as Artificial Neural Networks,Naive Bayes etc.which extracts knowledgeand features from voluminous of data is also popular in literature, (Beyer et al. 1994). Hybrid methods whichcombines different modelling approaches are also found in use to solve modelling and prediction of windspeed data, (Soman et al. 2010, Liu et al. 2014, Haque et al. 2013, Kiplangat et al. 2016, Drisya et al. 2017).

---

[1]Corresponding Author

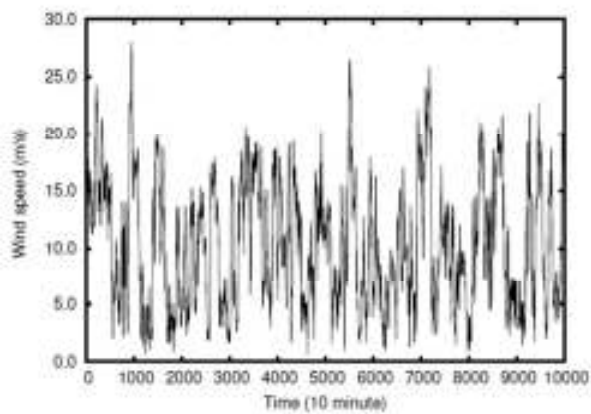

**Figure 1: Wind speed time series data measured at Latitude: 34:98420, Longitude: -104:03971 at 80 mheight.**

Since wind speed oscillations show distinct dynamical behaviour in different frequency components,(Drisya et al. 2017) in this work, we used wavelet based hybrid model for wind speed prediction. While, for the lower level high frequency component we opted AR method for modelling and prediction, higherlevel low frequency components were modelled using SVM. For the analysis, we used 10 min interval windspeed data measured at the period from January 2004 to December 2006 for the location Latitude: 34:98420,Longitude: -104:03971 at 80 metres height. The data is available at National Renewable Energy Laboratory (http://www.nrel.gov), USA.

## WAVELET TRANSFORM

Wavelet transform (WT), decompse the given compound signal into different time-frequency representationto analyse signals jointly in time and frequency. A prototype wavelet called mother wavelet is used as the basefunction which is then scaled and translated to get a series of wavelets for analysing temporal and frequencyvariations of the signal. The generated wavelet functions $\psi^{a,b}(t)$, from a mother wavelet $\psi$ is represented by

$$\psi^{a,b}(t) = \frac{1}{\sqrt{2}} \psi\left(\frac{t-b}{a}\right)$$

where $a \in \mathbb{R}\backslash 0, b \in \mathbb{R}$ and is expected to satisfy the admissibility condition, (Daubechies et al. 1992)

$$C_\varphi = \int_\mathbb{R} \frac{|\Psi(\omega)|^2}{|\omega|} d\omega < \infty$$

where $\Psi(\omega)$ is the Fourier transformation of $\psi(y)$.

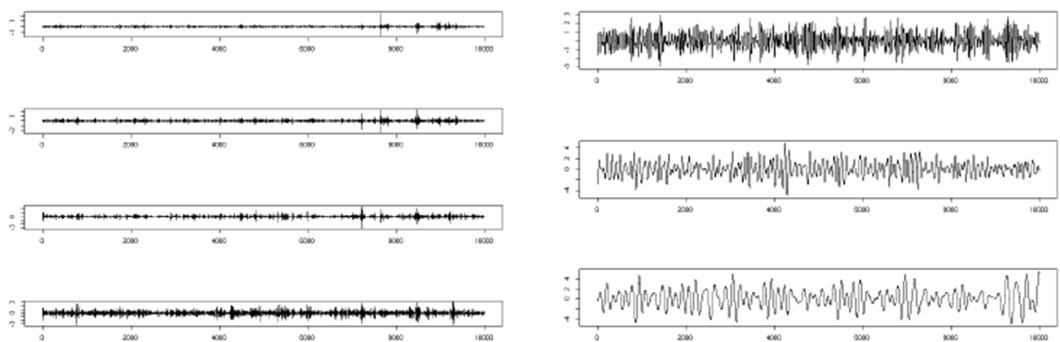

**Figure 2: Different wavelet components of the DWT of time series under consideration.(a) Shows thefrequency components from level 1 to 4 depicting the stochastic nature (b) the frequency components fromlevel 5 to 7 depicting the chaotic nature.**

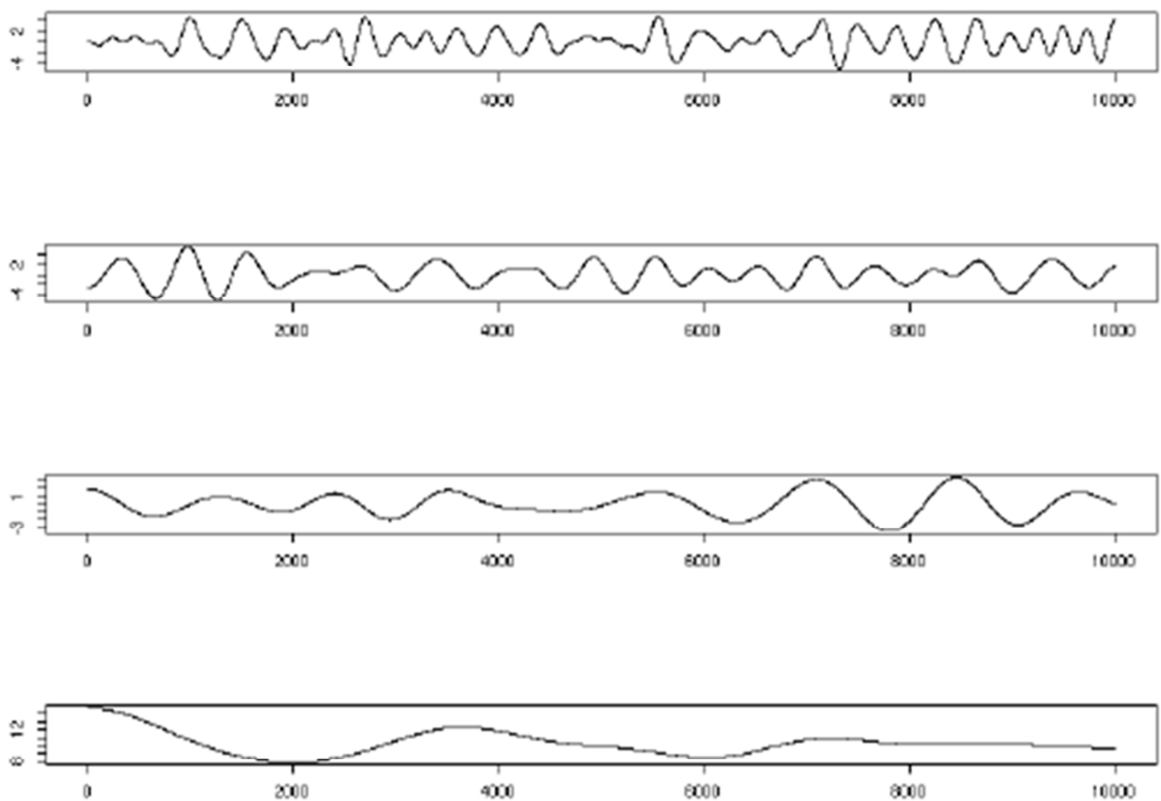

**Figure 3: Shows the frequency components from level 8 to 9 and the smooth component, depicting the deterministic nature.**

Discrete wavelet transform are more suitable for analysing time series data mesured at equal intervals of time and for suitably chosen dialation ($a_0 > 1$) and translation ($b_0$) parameters discrete wavelets are defined as

$$\psi(j,k) = a_0^{-j/2} \psi(a_0^{-j} t - k b_0), \qquad j, k \in \mathbb{Z}$$

The discrete wavelet transform(DWT) of signal $y(t)$ is then defined by,

$$DWT_{y(i,j)} = \int_{-\infty}^{\infty} y(t) \psi_{j,k}^*(t)\, dt$$

where $*$ denotes the complex conjugate. The admissibility condition of $\psi(t)$ allows the reconstruction of

$y(t)$ by the inverse transform, (Daubechies et al. 1992) and it is given by,

$$y(t) = \frac{1}{C_\psi} \sum_{j,k \in \mathbb{Z}} DWT_{y(j,k)} \psi_{j,k}(t)$$

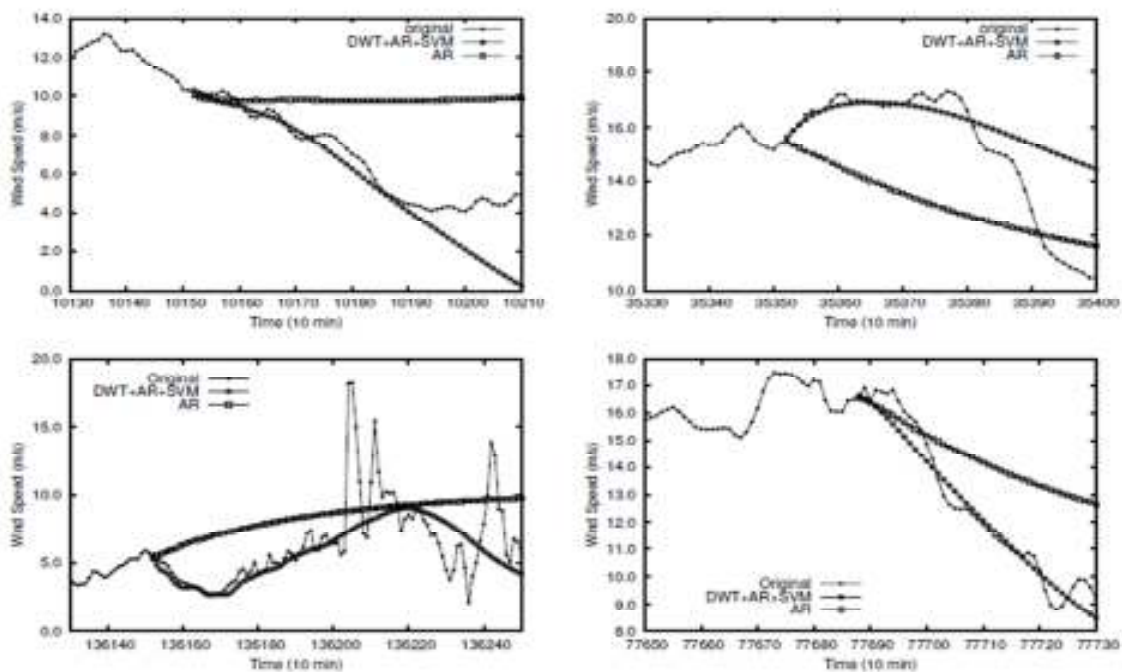

Figure 4: Comparison of predicted values with the actual values for DWT+AR+SVM and AR alone. Thesymbols are plotted only at every 30 minutes for clearness.

## SUPPORT VECTOR REGRESSION MODEL OF PREDICTION

Support vector machines (SVM) are a class of machine learning methods used for both classification and regression first introduced by Vladimir Vapnik. The basic idea of SVM is to construct the higher dimensional hyper planes from the given input data using kernal functions and choose the largest seperation betweenhyper planes to decide which class a new data point will be in, (Weston et al. 2001). The basic form of supportvector regression can be expressed as

$$z = \omega . \Phi(y) + b (\phi: R^n \to R^N)$$

where $y \in R^n$ is the input vector, $z$ is the target value, $b$ is a bias term and $\omega \in R^N$ is the coefficient vector. The nonlinear function which maps the input into higher dimension is represented by $\phi: R^n \to R^N$ and the vector $\Phi(y)$ can be of infinite dimension.

## AUTOREGRESSIVE PREDICTION MODEL

Based on the assumption that history of the state of a system has an effect on its current and future values, autoregressive(AR) model finds a linear combination of $d$ previous values as,

$$y_t = c + \sum_{i=1}^{d} \psi_i y_{t-i} + \epsilon_t$$

where $\psi_1, \psi_2, \ldots \ldots, \psi_d$ are the model coefficients, $c$ is the constant and $\epsilon_t$ is the Gaussian white noise, (Box et al. 2015). Parameter estimation of autoregressive model is done here with Burg's method, (De Hoon et al. 1996).

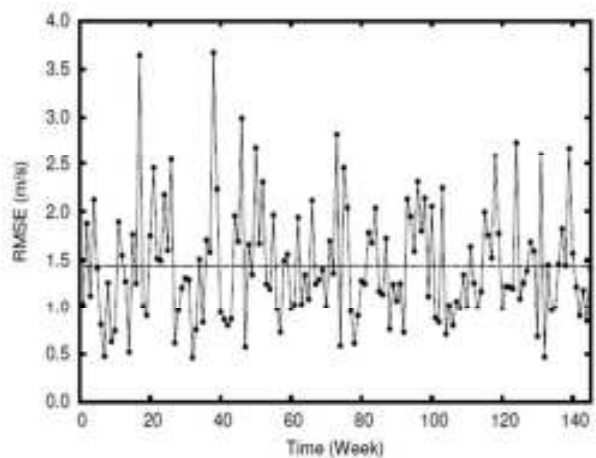

Figure 5: RMSE, with standard error, of 6 hour prediction calculate at every week for the period of 3 yearsfrom 2004 to 2006 for the location Latitude: 34.98420, Longitude: -104.03971.

## RESULTS AND DISCUSSION

Wind speed observations are known for its highly varying oscillations on very different time scales. A typical graphical representation of this highly variable wind speed fluctuations is shown in Fig.1. Our previous frequency level analysis of dynamical behaviour of wind speed oscillations, (Drisya et al. 2017) show that, at different scales the underlying dynamics can be classified as stochastic, chaotic and deterministic processes. The reported results indicate the suitability of different modelling techniques at each of the frequency components of the original time series. As a first step we applied the discrete wavelet transform with Daubechies wavelet on the measured time series data, to decompose it into 10 wavelet components, 9 detailed components and one smooth series. As seen in Fig.2(a) lower level frequency components levels up to 3 shows a complex behaviour suggesting the stochasting nature of the components. The complex dynamics seen in the intermediate levels of the decomposed series, Fig.2(b), is due its chaoticnature, (Drisya et al. 2017) and the very low frequency components depicted in Fig.3 are essentially deterministic innature. The stochastic process is modelled using AR method as given in Eq.7 assuming that there exist some amount of correlation with the previous values. Since the wavelet components beyond high frequency levels are deterministicin nature these signals are expected to be better predicted with deterministic models. Machine learning methods aresuitable for modelling deterministic dynamics because of their ability to capture the order present in the data. Fig.4 showsthe prediction made with simple AR method and the predictions made with the proposed hybrid model, AR techniquein high frequency wavelet components and SVM in low frequency level and it is evident that the results from hybridmethod improves the prediction accuracy significantly. For validating the results we extended our analysis and predictionto several data set from the same location and a root mean squared error (RMSE) is calculated. For a $k$-points forecast $y_{n+1}^p, y_{n+2}^p, \ldots\ldots, y_{n+k}^p$ the RMSE is calculated as,

$$RMSE = \sqrt{\frac{\sum_{i=n+1}^{n+k}(y_i - y_i^p)}{k}}$$

For the available data, 6 hrs ahead prediction is done at a 1 week interval for the 3-year period from 2004 to 2006. Timeseries data measured at the month January of the year 2004 were used for model building and all other predictions aredone with this constructed model. The RMSE values calculated are shown in Fig.5 and it is clear that for most of thepredictions upto 6 hr the RMSE is below 2:5 (m/s) and only a few prediction errors go beyond this. The average RMSEfor 6 hr prediction is shown as dotted lines in Fig.5 and it indicates that the proposed hybrid model can predict windspeed oscillations with an average error of 1:5 (m/s). Even though wind speed oscillations shows different dynamics indifferent time scale, suitable modelling techniques at each frequency level can significantly reduce the prediction error.

## CONCLUSION

Since the amount of power produced at given time is a direct function of the wind speed measured, accurate prediction of wind speed has got wider application in wind energy management. Most of the methods for predicting wind speed time series found in the literature are linear models and it could never accommodate the varying dynamical behaviour atdifferent time scale of the wind speed oscillations. In this work, we report that the use of a hybrid method, consideringthe varying dynamics at different levels and applying suitable modelling techniques, can remarkably improve forecast accuracy.

## ACKNOWLEDGEMENT

The authors are grateful to the Campus Computing Facility of the University of Kerala set up under DST-PURSE programmefor providing computational facilities. The first author, G. V. Drisya, would like to thank DST for the financialsupport through PURSE programme. Authors express their gratitude to the authorities of 27th swadeshi science congressfor giving the opportunity.